\documentclass[10pt, conference, compsocconf]{IEEEtran}
\IEEEoverridecommandlockouts
\usepackage{cite}
\usepackage{amsmath,amssymb,amsfonts}
\usepackage{textcomp}
\usepackage{xcolor}
\usepackage{algorithm}
\usepackage{algorithmic}
\usepackage{graphicx} 
\usepackage{hyperref}
\usepackage{subcaption}

\usepackage{dsfont}
\usepackage{bm}
\usepackage{float}
\usepackage{comment}
\usepackage{mathtools}
\usepackage{wrapfig}
\usepackage{lipsum}

\newcommand{\calig}[1]{\mathcal{#1}} 

\newcommand{\bo}[1]{\mathbf{#1}} 

\newcommand{\ba}{\begin{array}{rcll}} 
\newcommand{\ea}{\end{array}}

\newcommand{\ben}{\begin{enumerate}}
\newcommand{\bena}{\begin{enumerate}[(a)]}
\newcommand{\beni}{\begin{enumerate}[i)]}
\newcommand{\een}{\end{enumerate}}

\def\BibTeX{{\rm B\kern-.05em{\sc i\kern-.025em b}\kern-.08em
    T\kern-.1667em\lower.7ex\hbox{E}\kern-.125emX}}

\begin{document}

\title{Multi-level hypothesis testing for populations of heterogeneous networks$^1$}

\author{\IEEEauthorblockN{Guilherme Gomes}
\IEEEauthorblockA{Department of Statistics\\
Purdue University\\
West Lafayette, USA \\
gomesg@purdue.edu}
\and
\IEEEauthorblockN{Vinayak Rao}
\IEEEauthorblockA{Department of Statistics \\
Purdue University\\
West Lafayette, USA \\
varao@purdue.edu}
\and
\IEEEauthorblockN{Jennifer Neville}
\IEEEauthorblockA{Department of Computer Science \\
Purdue University\\
West Lafayette, USA \\
neville@cs.purdue.edu}
}

\maketitle

\footnotetext[1]{This is the extended version of \cite{gomes2018multiple}.}
\begin{abstract}
In this work, we consider hypothesis testing and anomaly detection on datasets where each observation is a weighted network. Examples of such data include brain connectivity networks from fMRI flow data, or word co-occurrence counts for populations of individuals. Current approaches to hypothesis testing for weighted networks typically requires thresholding the edge-weights, to transform the data to binary networks. This results in a loss of information, and outcomes are sensitivity to choice of threshold levels. Our work avoids this, and we consider weighted-graph observations in two situations, 1) where each graph belongs to one of two populations, and 2) where {\em entities} belong to one of two populations, with each entity possessing multiple graphs (indexed e.g.\ by time). Specifically, we propose a hierarchical Bayesian hypothesis testing framework that models each population with a mixture of latent space models for weighted networks, and then tests populations of networks for differences in distribution over components. Our framework is capable of population-level, entity-specific, as well as edge-specific hypothesis testing. We apply it to synthetic data and three real-world datasets: two social media datasets involving word co-occurrences from discussions on Twitter of the political unrest in Brazil, and on Instagram concerning Attention Deficit Hyperactivity Disorder (ADHD) and depression drugs, and one medical dataset involving fMRI brain-scans of human subjects. The results show that our proposed method has lower Type I error and higher statistical power compared to alternatives that need to threshold the edge weights. Moreover, they show our proposed method is better suited to deal with highly heterogeneous datasets.
\end{abstract}
\section{Introduction}
\label{sec:intro}

\noindent We consider the problem of graph-based hypothesis testing, which tests whether
two sets of graph-valued observation
samples are drawn from the same distribution. This is a topic of 
growing interest~\cite{li2011graph,asta2015geometric,muk2017,arrigo2017exponential};
however, there are only a few studies 
where the observations are {\em weighted graphs}. In this work, we address 
this gap, considering the problem of {\em hypothesis-testing for 
  replicated weighted graph-valued data}. 
This is a challenging problem, since the average and atypical behavior 
of a sample of networks
is difficult to characterize. 

\begin{figure}
\centering
 \hspace{-12mm}
\includegraphics[width=.55\textwidth]{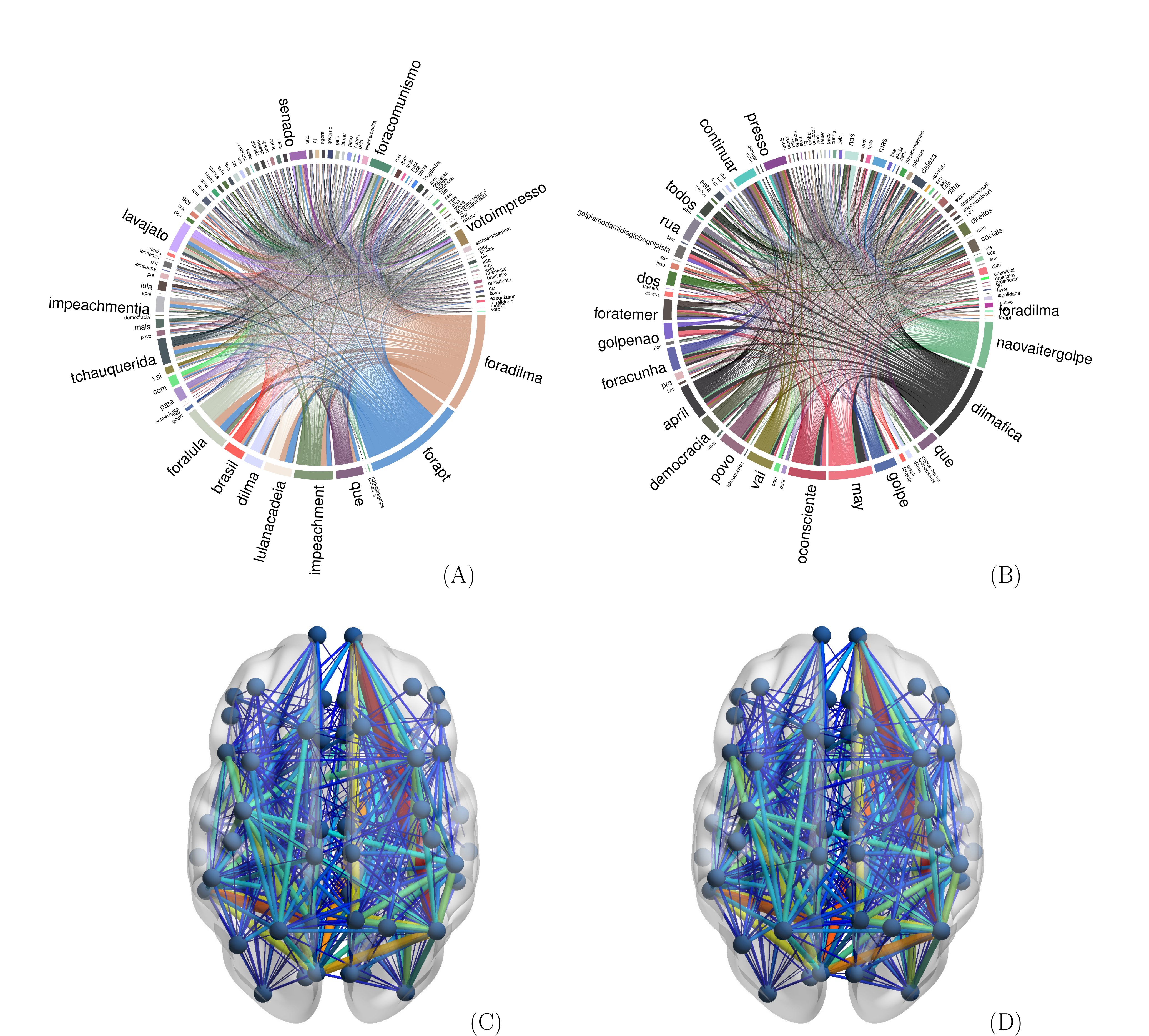} 
\caption{Top row: word connectivity networks for two Brazilian Twitter users, (A) from the pro-government side and (B) from the anti-government side. Bottom row: brain connectivity 
network for two individuals, (C) female and (D) male.}
\label{fig:net_twitter} 
\end{figure}

Figure \ref{fig:net_twitter} illustrates two example domains with populations of weighted graphs. The top row illustrates the word co-occurrence networks of two Twitter users in Brazil, one that is {\em pro-government} and one that is {\em anti-government}. Here the entity corresponds to a user on social media,  the nodes in the graph are vocabulary words, and the edges weights reflect co-occurrences of words in the posts of the users. 
The bottom row illustrates brain connectivity networks of two individuals, one {\em female} and one {\em male}. Here the entity corresponds to an individual, the nodes in the graph are brain regions, and the edges weights represent functional connectivity strength as measured by functional magnetic resonance imaging (fMRI).
In both cases, we would like to investigate how populations and entities differ. For example, whether brain activity differs with respect to sex or whether word usage differs with respect to political views.

There has been recent work on graph-based hypothesis testing (see \cite{akoglu2015graph} for a good survey). However, much of the work has focused on testing subgraphs {\em within} a larger graph (e.g., \cite{mongiovi}), or on one-sample tests comparing a single graph to a null model (e.g., \cite{moreno2}). Work focusing on {\em populations} of graphs has received considerably less attention 
and falls into one of two categories: that 
of~\cite{ginestet2014}, which introduces a geometric characterization of 
the network  using the so-called Fr\'echet mean, and that 
of~\cite{durante2016}, who proposed a Bayesian latent-variable model for
unweighted graphs.  We focus on the latter, which allows us to bring the 
powerful machinery of probabilistic hierarchical modeling to the table, 
allowing noisiness and missingness, and providing interpretable confidence 
scores. Unfortunately, existing work along this second direction is limited 
to modeling binary graphs, so that in practice, a {\em threshold} 
must be used to transform counts or continuous weights to 0/1 values.
Such a thresholding operation discards valuable information about 
the strength of the edge-weights, and can also exhibit sensitivity towards 
the choice of threshold. Too small a threshold can result in a graph that 
is too dense, and too large, too sparse. Often, there does not even 
exist a single appropriate threshold across the entire graph.

In this paper, we address the issues of previous work and develop a hypothesis testing framework that facilitates testing over graphs populations with 
edge-weights, which can follow any parametric distribution. 
Specifically, we propose a Bayesian hypothesis testing framework that uses a mixture of 
latent space models for weighted networks to test for population-differences.
Our framework is capable of population-level, entity-specific as well 
as edge-specific hypothesis testing. 
We consider testing 
in two broad scenarios: 
\begin{enumerate}
  \item When all {\em observations} from the same population follow the same distribution, we can ask:
    Are the population distributions identical?
  \item When all {\em entities} from the same population follow the same distribution, we can ask: 
    Are the population distributions identical? Now, every entity has an 
    associated set of graph-valued observations which are identically 
    distributed, but are not exchangeable across 
    entities.
\end{enumerate}
    Observe the first case is equivalent to the second when each 
    entity has only one associated graph; however the latter allows 
    heterogeneity among entities in the same population. For instance an 
    entity in a population that is politically  {\em conservative} might frame an issue they discuss from an economic 
    perspective, while another entity in a population that is politically  {\em liberal} might focus on the social aspects of the same issue.  \cite{peixoto2015inferring} 
    proposed a strongly parametric time-varying framework to handle this 
    important situation, our approach is significantly more flexible.

We apply our testing framework to problems from the types of domains summarized in
Figure \ref{fig:net_twitter}.
First, we look at word co-occurrence network data from Twitter (on the 
political crisis in Brazil), as well as Instagram (on side 
effects of Adderall and Ritalin usage for Attention Deficit Hyperactivity 
Disorder~\cite{correia}). In both datasets, 
we investigate how populations and entities differ 
based on the way they communicate---specially in the manner in which the usage of pairs of key words differs between groups. 
Standard methods such as unigram mixture models, latent Dirichlet  
allocation (LDA)~\cite{blei2003latent} or N-gram language  
models~\cite{katz1987estimation}, which are based just on word-frequency,
do not capture the kind of contextual information we are interested in. 
While these methods can identify words that `belong' to different groups, in our scenario there is a strong overlap in key words across groups, 
and such models will fail to differentiate between groups which share common themes and 
vocabularies. 
Second, we used functional magnetic resonance imaging (fMRI) 
data,
to investigate how brain activity differs across populations like sex, age and 
personality traits like extroversion, conscientiousness and creativity. 
In both tasks, 
we show that graph-structure as well as graph weights are crucial to 
performance, and that we outperform baselines like latent Dirichlet allocation 
(LDA)~\cite{blei2003latent}, N-gram language 
models~\cite{katz1987estimation}, as well as thresholding methods 
like~\cite{durante2016}.

\vspace{2mm}
\noindent \textbf{Contributions:}
Our contribution is a multi-level statistical hypothesis testing framework for populations of weighted networks, 
concerning both the overall graph distributions, as well as two types of local 
hypotheses: 
entity-specific and edge-specific. The latter are important since a population
might have networks, or a network edges, that are 
statistically different, and that might escape detection by a global test. 
Our hierarchical Bayesian mixed membership model allows statistical
information to be shared across groups, increasing 
accuracy of hypothesis tests without loss of statistical power. This allows practitioners to evaluate anomalies 
in a principled manner, using statistical significance. Notably, our framework is more robust than previous methods developed for binary graphs, which require thresholding of weighted data before application. 

\section{The model}
We are given a set $\bo{A}$ of undirected graphs, with graph $A_{nt}$ 
belonging to entity $n$ at index $t$ (refered to as `time'). Here 
$n \in [1,...,N]$  and $t \in [1,...,T_n]$, with $A_{nt}[i,j]$ giving the 
link strength between vertices $i$ and $j$ of entity $n$ at time $t$ 
($i,j \in [1,...,V]$).  
We also observe population information $y_{n} \in [1,...,G]$ for each entity. 
For instance, each network might represent word co-occurrences in a user's 
social media messages over some time period, while the population might 
indicate whether the user's political leanings are `Liberal' or 
`Conservative'. 

Underlying our testing framework is a probabilistic model which we now 
outline. We assume each observation $A_{nt}$ comes from one of $H$ clusters or 
mixture components, with cluster $h$ having parameter $\bm{\theta}^{(h)}$.
Each cluster has a distribution over graphs which we write as 
$F(\bm{\theta}^{(h)})$ (we specify $\bm{\theta}^{(h)}$ and $F$ in the next paragraph).
Clusters and cluster parameters are shared across populations, however each 
population $y$ has its own Dirichlet-distributed probability over clusters, 
$\bm{\beta}_y$. At a high-level ours is a Bayesian hypothesis-testing 
approach which tests whether the $\bm{\beta}_y$'s are identical across 
populations. For the case of two populations, we place equal {\em a priori} 
probability on the null hypothesis $H_0: \bm{\beta}_1 = \bm{\beta}_2$ and 
the alternative $H_1: \bm{\beta}_1 \ne \bm{\beta}_2$. Using the machinery 
of Bayesian inference, 
we evaluate the posterior probabilities of the two hypotheses given 
observations, and reject the null if its probability $P(H_0 | -)$ is less than some 
specified threshold (e.g., 0.05 or 0.1). 

We now describe the cluster-specific distribution over graphs,
$F(\bm{\theta}^{(h)})$. For cluster $h$, $\bm{\theta}^{(h)}$ is a $V\times V$
matrix, whose $(ij)$th element parametrizes the probability of the weight 
on the edge between nodes $i$ and $j$. In our applications, we looked at 
count-valued edges, and so assumed $F\left(\bm{\theta}^{(h)}\right)$ 
to be Binomial or Poisson distributed with parameter $\bm{\theta}^{(h)}[i,j]$
on edge $(i,j)$.
We define $\bm{\theta}^{(h)} = f\left(\bm{S}^{(h)}\right)$ where $f(\cdot)$ 
is some link function (e.g.\ the logistic or exponential function to 
ensure nonnegativity), and 
constrain $\bm{S}$ using a low-rank factorization scheme $\bm{S}^{(h)} = \bo{X}^{(h)}
\bo{X}^{(h)T}$. Here $\bo{X}^{(h)} \in \mathcal{R}^{|V| \times R}$ and 
$R \ll |V|$, so that $\bo{X}_v^{(h)}$ gives the location of node $v$ 
in some low-dimension space, and $\bm{S}^{(h)}$ is the proximity of all 
nodes. The number of parameters thus grows linearly, rather than quadratically
with the number of vertices. In equations, we expand the upper plates in 
Figure~\ref{mixGraph}(a) and (b), to get
\begin{align}
\bm{\theta}^{(h)}& = f(\bm{X}^{(h)}\bm{X}^{(h)T}), \quad 
\bo{X}_{v}^{(h)} \sim N_{R}\left(\bm{0},\mathds{I}\right), v=1\ldots,V 
\end{align}
Each population $y$ has a distribution over clusters $\bm{\beta}_y$. With 
prior probability half, the null hypothesis is true 
(we indicate this with the variable $\mathcal{T}$), in which case all 
populations have the same distribution $\bm{\beta}$. Otherwise, each population $g$
has its own distribution, $\bm{\beta}_g$.
 Thus,
\begin{align}
  \mathcal{T} \sim  \text{Bern}&(1/2) \\
  \text{If } \mathcal{T} = 0:& \quad
  \bm{\beta}_1 =\ldots,=\bm{\beta}_G \sim \text{Dir}(\alpha,...,\alpha) \nonumber \\
\text{Else:} &\quad
\bm{\beta}_g \stackrel{iid}{\sim} \text{Dir}(\alpha,...,\alpha) \text{ for } g=1\ldots G.\nonumber 
\end{align}

Now, consider the case where each 
entity has only a single associated graph. Then the $n$th entity
(belonging to population $Y_n$) has a graph $A_n$ distributed as
\begin{align}
 C_{n}|Y_{n}  \sim \bm{\beta}_{Y_n}  \quad 
\bo{A}_{n}|C_{n} \stackrel{}{\sim} F\left(\bm{\theta}^{(C_n)}\right)
\label{eq:gen_model0}
\end{align}
Here $C_n$ refers to the latent variable that identifies the cluster membership of entity $n$, which depends on the population entity $n$ is drawn from.

For the case where we have multiple network observations per entity, we add a 
layer to this hierarchical model. 
Now, each entity $n$ has their own distribution over clusters 
$\bm{\pi}_n$ centered around the population distribution:
\begin{align}
  \bm{\pi}_{n} \sim Dir(\beta_{Y_n}). 
\end{align}
The graph $t$ of this entity is independently distributed as
\begin{align}
C_{nt}|Y_{n}  \sim \bm{\pi}_{n},\quad 
\bo{A}_{nt}|C_{nt}  \stackrel{}{\sim} F\left(\bm{\theta}^{(C_{nt})}\right)
\label{eq:gen_model}
\end{align}
Figure (\ref{mixGraph}) summarizes our generative 
process for both cases.
\subsection{Model Inference}
We are given a set of network observations ${\bm A}$, each written as 
$A_{nt}$ where $n$ indexes entities and $t$, `time'.
For each $A_{nt}$, we are also given 
a population assignment $Y_n\in \{1,2\}$. 
Since we observe the population memberships $\bm{Y}$ and the 
networks $\bm{A}$, the inferential task is to learn 
$C_{nt}$, ${\bm\pi}_{n}$, $\bm{\beta}_y$ and $\bm{\theta}^{(h)}$.  
In the next section, we will use these variables as statistics in our hypothesis tests.
For notational convenience, we will refer to a link between an arbitrary 
pair of nodes $i$ and $j$ with $l$, so that we can write $A_{n}[i,j]$ as 
$A_{n}[l]$. 
We also represent the weighted matrix with its vectorized 
lower triangular component  $\mathcal{L}(\bo{A}_n) = 
(\mathcal{L}(A_{nt})_1,...,\mathcal{L}(A_{nt})_{V(V-1)/2})$.  
For the general model specified above, we carry out posterior inference 
via a Gibbs sampler, whose individual updates we outline next:

\begin{enumerate}
  \item {\em Sample the cluster indicator for each graph.} \\This comes from the multinomial:
\begin{multline*}
 P(C_{nt}=h|- ) =\\ \frac{\pi_{n_h} \beta_{h} \prod_l P\left( \calig{L}(A_{nt})_l = a_l|\theta_{l}^{(h)}\right)}{\sum_{m=1}^H \pi_{n_m} \beta_{m} \prod_l P\left( \calig{L}(A_{nt})_l = a_l |\theta_{l}^{(m)}\right)}
\end{multline*}

  where $l \in [1,...,V(V-1)/2]$ and $a_l|\theta_l^{(h)} \sim \left[F(\bm{\theta}^{(h)})\right]_l$.
\item {\em Sample the mixing probabilities for each entity $n$.} \\
  With $\bm{{m}}_n$ a vector of cluster assignment counts of graphs of 
  entity $n$
$$
\bm{\pi}_n \sim \text{Dir}(\bm{\beta}_{Y_n} + \bm{{m}}_n)
$$
\item {\em Sample the locations of the nodes for each cluster.} \\With a Gaussian 
  prior over the locations $\bm{X}_n$, and the weight-distribution parametrized after 
  transforming through a 
  link function $f$, this is a straightforward exercise in sampling 
  from the posterior of a Gaussian with a nonlinear link function. Standard
  techniques exist to do this~\cite{duane1987hybrid,ellslisam}, though we 
  followed a recent idea involving the Polya-Gamma data-augmentation 
  scheme~\cite{polson2013}.
\item {\em Sample the testing indicator $\mathcal{T}\sim \text{Bern}(P(\mathcal{H}_1)|-)$.}
  Since $\mathcal{T}$ is the test-statistic central to our methodology, we discuss 
  this in a bit more detail in the next section. The update rule is 
  given by Equation~\ref{probab}.
\item {\em Sample the mixing probabilities for each population $y$.} \\
If $\mathcal{T}=0$ then for all $y$, $\bm{\beta}_y = \bm{\beta} 
\sim \text{Dir}(\alpha+n_1,...,\alpha+n_H)$ where $n_h$ is the number of graphs in 
cluster $h$.
If $\mathcal{T}=1$ then $\bm{\beta}_y \sim \text{Dir}(\alpha+n_{1y},...,\alpha+n_{Hy})$ for each 
$y$, where $n_{iy}$ counts the number of graphs from population $y$ in 
cluster $i$.
\end{enumerate}

\section{Weighted-network comparison tests}
\label{sec:datagen}

For simplicity, we focus on the case where we have only two populations ($G=2$).
Under our formulation, the problem of hypothesis testing 
amounts to testing whether the population-level cluster assignment probabilities 
$\bm{\beta}_1$ and $\bm{\beta}_2$ are significantly different under the 
posterior. We elaborate on this below.

\begin{figure}
\hspace{-8mm}\includegraphics[width=0.6\textwidth]{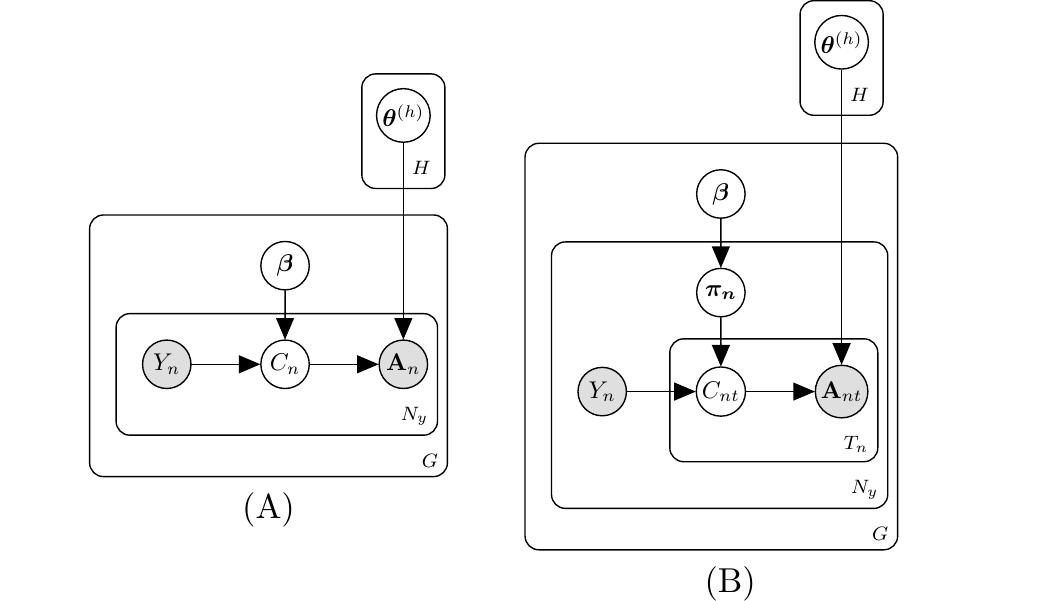}
\caption{The graphical models are given by (A) fixed and (B) with 
time-varying structures. }
\label{mixGraph}
\end{figure}

\subsection{Population-level network comparison test}
This task involves comparing the posterior probabilities of the 
two hypotheses, $H_0: \bm{\beta}_1 = 
\bm{\beta}_2$ vs $H_1:\bm{\beta}_1\ne \bm{\beta}_2$. Since $H_0$ being true 
amounts to $\mathcal{T}=1$, our MCMC estimate of the probability equals 
the fraction of MCMC iterations where $\mathcal{T}=1$. We first describe how our 
Gibbs sampler updates this variable (step 4 of our Gibbs sampler).
At any MCMC iteration, let $\bm{{m}}_y$ be the vector of 
cluster assignment counts for population $y$, with component $c$ giving the number 
of observations from population $y$ assigned to cluster $c$:
$\bm{{m}}_y = \left(\sum_{n;y_{n}=y}\sum_t \mathds{I}_{C_{nt}=1},...,\sum_{n;y_{n}=y}\sum_t \mathds{I}_{C_{nt}=H}\right)$. 
We write $\bm{{m}}=\bm{{m}}_1 + \bm{{m}}_2$ (for the two populations in $G$, i.e., 1 and 2). 
Then, under the two hypotheses, these counts are distributed as
\begin{equation}
\begin{aligned}
  \mathcal{H}_0:& \ \bm{{m}}_1, \bm{{m}}_2 \stackrel{iid}{\sim} \text{Mult}(\bm{\beta}),  \\
              & \ \bm{\beta} \sim \text{Dirichlet}(\bm{\alpha}) \\
  \mathcal{H}_1:& \ \bm{{m}}_1 \sim \text{Mult}(\bm{\beta}_1) \ \ and \ \ \bm{{m}}_2 \sim \text{Mult}(\bm{\beta}_2) \\
              & \ \bm{\beta_1},\bm{\beta_2} \sim \text{Dirichlet}(\bm{\alpha}) 
\end{aligned}
\label{overall_hyp}
\end{equation}
Marginalizing out the $\bm{\beta}'s$, and recalling that both hypotheses have 
the same prior probability, we can specify the posterior
$$P(\mathcal{H}_1|-) = \frac{P(\bm{{m}}_1|\bm{\alpha})P(\bm{{m}}_2|\bm{\alpha})}{P(\bm{{m}}|\bm{\alpha})+P(\bm{{m}}_1|\bm{\alpha})P(\bm{{m}}_2|\bm{\alpha})}$$
From the Dirichlet-multinomial conjugacy, we can write down the
marginal probabilities of the observations, giving
\begin{equation}
P(\mathcal{H}_1|-) = \frac{\prod_{y=1}^2 \frac{B(\bm{\alpha}+\bm{{m}}_y)}{B(\bm{\alpha})}}{\frac{B(\bm{\alpha}+ \bm{{m}})}{B(\bm{\alpha})}+ \prod_{y=1}^2 \frac{B(\bm{\alpha}+\bm{{m}}_y)}{B(\bm{\alpha})}}
\label{probab}
\end{equation}
where $B(\cdot)$ is the multivariate beta function $B(x) = \prod_{i=1}^q\frac{\Gamma(x_i)}{\Gamma(\sum_{i=1}^q x_i)}$.  
Every Gibbs iteration samples $\mathcal{T}$ from this, with the posterior probability 
of the alternative hypothesis, $P(\mathcal{H}_1|-)$, being the 
fraction of MCMC samples where $\mathcal{T}$ equals $1$. If the estimate 
from Equation~\eqref{probab} is larger 
than a specified threshold {(e.g., 0.95)}, we reject the null hypothesis and conclude that the populations are significantly different. We can use this {\em network comparison} (NC) test for both models in Figure~\ref{mixGraph}. When we use the (fixed) model in \ref{mixGraph}a, we will refer to it as NC-F and when we use the (mixed-membership) model in \ref{mixGraph}b, we will refer to it as NC-M.

\subsection{Entity-specific comparison test}
This task refers to do following hypothesis test: $H_0^{n_1n_2}: \bm{\pi}_{n_1} = 
\bm{\pi}_{n_2}$ Vs $H^{n_1n_2}_1:\bm{\pi}_{n_1} \ne \bm{\pi}_{n_2}$
for any two users $n_1$ and $n_2$.
Estimating this from our posterior samples is straightforward. Assuming 
multiple networks per entity, let 
$\bm{\tilde{m}}_{n}$ be a vector of counts for entity $n$, giving the number 
of observations assigned to each cluster. As mentioned earlier, the
entity-specific distribution over clusters follows the distribution 
$\bm{\pi}_n \sim \text{Dir}(\bm{\beta})$. Following the earlier logic, Equation~\ref{local_hyp} gives the posterior probability two given entities have
different cluster assignment probabilities:

\begin{equation}
P(\mathcal{H}^{n_1n_2}_{1}|-) = \frac{\prod_{i=1}^2 \frac{B(\bm{\beta}+\bm{\tilde{m}}_{n_i})}{B(\bm{\beta})}}{\frac{B(\bm{\beta}+ \bm{\tilde{m}})}{B(\bm{\beta})}+ \prod_{i=1}^2 \frac{B(\bm{\beta}+\bm{\tilde{m}}_{n_i})}{B(\bm{\beta})}}
\label{local_hyp}
\end{equation}
It is important to note that Equation \ref{local_hyp} allows pairwise 
comparisons across populations, and therefore it is possible to have 
significantly similar entities from different populations and significantly 
different entities in the same population.

\subsection{Edge-specific comparison test}
This task refers to the following hypothesis test for an edge $l=(i,j)$,
$H_0^l: \bm{{\theta}}_1[l] = \bm{{\theta}}_2[l]$ vs $H_1^l: \bm{{\theta}}_1[l] \ne \bm{{\theta}}_2[l]$.
For the edge application we use an adjusted version of Cramer's V-statistic proposed by~\cite{dunson2009} given by Equation~\ref{edgeesp}:

\begin{equation}
p_l^2 = \sum_{y=1}^2 \mathbf{p}_{\mathcal{Y}} \sum_{a_l} \frac{ P(\calig{L}(A)_l = a_l|\bar{\theta}_{y_l}) -  P(\calig{L}(A)_l = a_l|\bar{\theta}_{l})}{ P(\calig{L}(A)_l = a_l|\bar{\theta}_{l})}
\label{edgeesp}
\end{equation}
 where $\mathbf{p}_{\mathcal{Y}}$ is the sample size proportion of each population, $\bm{\bar{\theta}}_y =\sum_{n=1}^{N} \sum_{h=1}^H \beta_{yh} \bm{\theta}^{(h)}   \mathds{I}_{y_n=y} $, and $\bm{\bar{\theta}} = \sum_{y=1}^2\frac{\bm{\bar{\theta}}_y}{2}$.
If $p_l \approx 1$ then there is  evidence that edge weights differ across the populations.

\section{Related work}
\label{sec:relwork}

Graph-based hypothesis testing and anomaly detection are 
topics of growing interest with diverse applications~(see e.g., \cite{akoglu2015graph}). Many applications of hypothesis testing in network analysis focus on subgraphs {\em within} a larger graph (e.g., \cite{mongiovi}), or one-sample tests comparing a single graph to a null model (e.g., \cite{moreno2}). 

Work on populations of graphs 
can be divided on two areas: dynamic networks, in which one graph is
replicated over time \cite{sarkar2012nonparametric}, 
\cite{charlin2015dynamic} and \cite{durante2014nonparametric}; and 
exchangeable graph modeling in which each graph is considered to be one 
observation for a single entity (see~\cite{ginestet2014,asta2015geometric,durante2015,durante2016} 
and~\cite{duvenaud2015convolutional}). 

Our paper generalizes work from the latter category by allowing within-population
heterogeneity, with each entity having multiple graphs with similar
statistical properties.  Both~\cite{ginestet2014} and~\cite{asta2015geometric} 
deal with geometric characterizations of networks, and while their 
approaches are mathematically elegant, they are substantially less 
flexible than our work. \cite{duvenaud2015convolutional} take a 
convolution neural network approach for non-aligned graphs, where there 
is no known mapping between nodes in each graph.
This, coupled with the fact that their method requires the presense of 
node features, makes it unsuitable for our applications.  

Most closely related to ours is the method presented in \cite{durante2016}.
This method, which we will refer to DD, is a special case of our framework, 
where there is no within population variation, and where network edges 
are binary.  For count or continuous-valued data, one might consider 
thresholding the edge weights of each entity and then applying the DD 
method.  However this discards valuable information about the strength of 
the edge-weights, introduces sensitivity to threshold-level, and can 
reduce statistical power. Our method offers the ability to flexibly model 
such edge-weight information without any significant additional 
computational complexity. In particular, the computational time 
complexity of both our method and DD is $\mathcal{O}(NHR|V|^2)$ per 
iteration. {Here $N$ is the number of networks, $H$ is the number of 
clusters, $R$ is the dimensionality of the low rank approximation, and $|V|$ is 
the number of nodes in each graph. In practice, $H$ and $R$ are small 
constants that do not grow with the data.} 
In our experiments we compare with DD for different thresholds.
\section{Experiments}
\label{sec:syndata}

\begin{figure*}
 \hspace{-8mm}
\centering
\includegraphics[width=1.04\textwidth]{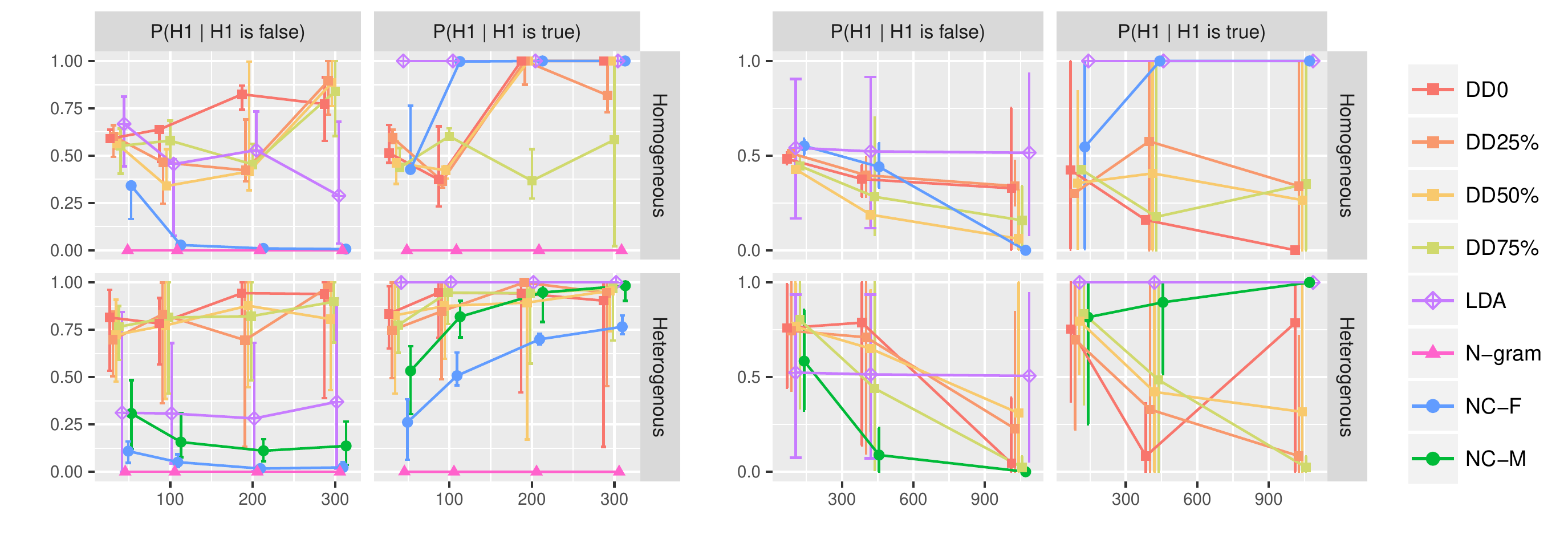} 
 \caption{Type I error and statistical power curves for the synthetic (left) and twitter (right) data for increasing sample sizes}
 \vspace{-4mm}
\label{fig:syn_power}
\end{figure*}

In order to assess the efficacy of our method, we divided our analysis into 
four parts: statistical power and type-I error  analyses, 
population-level hypothesis tests, edge-specific hypothesis tests, and 
additional exploratory analysis. 

We start with statistical power and type-I error analyses, the 
most important measures of assessing hypothesis tests. 
We investigate the efficacy of our (and competing) methods for varying 
sample sizes when the ground truth in known. We show that when the data 
are generated from a known two-population setup, our hypothesis testing 
framework produces significantly more accurate results and has lower 
variance, with respect to type-I error and statistical power, compared 
to a number of other baselines.  We show that for time-varying data, the 
mixed membership extension of our model is essential for reliable 
inference.  We also study the sensitivity of the method of~\cite{durante2016} 
(which requires unweighted graphs) to threshold settings, for 
population-level hypothesis tests. We show that for heterogeneous data, 
the hypothesis-test decisions are highly sensitive to threshold choice. 
We study the edge-specific hypothesis tests qualitatively, by visualizing 
the estimated model structure for each approach. We end by describing 
some additional insights that that our method gleans from the 
data. We start by describing the datasets.

\subsection{Datasets}
We generated synthetic weighted network data for two settings, the 
entity-homogeneous version from Figure~\ref{mixGraph}a (NC-F, where each 
entity is represented by one graph) and the entity-heterogeneous version 
from Figure \ref{mixGraph}b (NC-M, where each entity is represented by 
multiple graphs). We also applied our framework to three real-world 
applications: a Twitter dataset from the political crisis in Brazil, two 
datasets about drugs usage on Instagram, and fMRI recordings of brains of 
human subjects.

\vspace{1mm}
\noindent \textbf{Synthetic data (Homogeneous):}
\label{syn:overlap}
We generate synthetic data from two populations whose underlying weight 
probability matrices ${\bm \theta}$ overlap around the middle 
set of nodes, but where population $1$ has an elevated pattern of weight 
values for the first set of nodes, and population $2$ in the final set.  
Figure \ref{struc3} in Appendix \ref{sec:struc} shows this structure. 
We simulated 200 entities per population, with 100 nodes for each network. 
Given the structure, the weights of the edges were distributed according to a multivariate Zipf distribution \cite{paretomulti2002}. 
See Appendix section \ref{sec:struc} for more details. 

\vspace{1mm}
\noindent \textbf{Synthetic data (Heteromogeneous):}
\label{syn:time}
  Using the same population structure as above, we also construct a time-varying 
  dataset where each individual has four time points, resulting in four 
  different graphs per entity. In this case, we have 50 entities and 100 
  nodes for each network. Given the dependency structure, the weights for 
  each entity at each time point were distributed according to a 
  multivariate Zipf distribution. 
  Figure \ref{struc_time} in Appendix \ref{sec:struc} shows these structures.
  We use this dataset to compare the 
  behavior of the NC-F and NC-M models under different scenarios. 

\vspace{1mm}
\noindent \textbf{Real data:}

\noindent {\em Twitter:}
Brazil has recently faced the worst economic/political crisis of its republic years.
People were largely split into two sides: one who argued for the impeachment 
of the now former president, Dilma Rousseff, and the opposition who 
claimed that the process was a government coup. We crawled public Twitter posts 
from April 6th to May 31st 2016, using hashtags from both sides to 
collect tweets. The resulting dataset consists of $7,447$ users (entities), 
$4,233$ for the proposition and $3,214$ for opposition. In order to have 
appropriate data for the heterogeneous setting, we also split the dataset 
into time intervals, with each user having a network for every two weeks 
of tweets. We call this dataset ``Twitter time''. In this dataset, 
consisting of users with at least 15 days of tweets, we have a total of 
$2,098$ users, $1,255$ for proposition and $843$ for opposition. 
Figure \ref{fig:net_twitter} shows sample co-occurrence networks from the 
two sides: Proposition and Opposition. Each edge-weight indicates the 
number of tweets of a user $n$ containing two words (nodes) in a time 
interval $t$.  

\noindent {\em Instagram:}
We collected public Instagram comments with hashtags referring to the two most common  drugs to treat ADHD (Adderall and Ritalin) and Depression (Prozac and Zoloft). These medications all have additional uses (and consequently symptoms), for instance, Adderall is known for loss of  appetite, and as an aid for academic performance.  Our dataset consists of $65$ users with $44,408$ posts for \#adderall, $21$ with $17,466$ for \#ritalin, $111 $ with $129,405$ for \#prozac, and $35$ with $29,357$ for \#zoloft. 

\noindent {\em fMRI brain images:}
Functional magnetic resonance imaging (fMRI) captures 
activity in the brain by measuring blood flow from one region of the 
brain to another. We used the MRN-111 dataset\footnote{\url{http://openconnecto.me/data/public/MR/}} which consists of functional 
magnetic resonance images (fMRI) for 114 subjects (entities). 
As in \cite{desikan2006automated} we used a total of 68 brain regions, 34 from the left hemisphere and 34 from the right. Nodes represent brain regions, and weights, white matter density across nodes. We compare brain activity across characteristics like 
Sex ($Male$ vs $Female$), and personality traits like creative level 
($\le 90$ vs  $\ge 111$), extroversion ($\le 30$ vs  $\ge 35$). Values 
for creative level (CCI) and  extroversion are given from a psychometric 
scale determined by the corresponding scientific literature, those thresholds 
were chosen to illustrate a clear \textit{LowVsHigh} setting. 
Figure \ref{fig:net_twitter} shows sample brain networks of the MRN111 dataset for female and male individuals. We observe significant variability in these weights, suggesting that thresholding can lead to loss of information. 

\subsection{Baselines}

We compared our NC-F and NC-M methods with the following baselines:
\begin{enumerate}
\item LDA (topic modeling)~\cite{blei2003latent}: This treats each entity as 
a document made up of `topics' (each corresponding to a distribution
over word-count patterns).
\item N-gram language model \cite{katz1987estimation}: We use 
observed bigrams frequencies to estimate co-occurrence probabilities.
\item DD network model~\cite{durante2016}: As stated previously, DD  
forms a special instance of our more general framework for unweighted networks. 
In order to apply DD, we need
first threshold the weighted network observations. We do so using the 
following criterion \cite{simas2015distance}:  
$$ p_{ij} =\frac{\text{co-occurances between words $i$  and $j$}}
{\text{min (counts of words $i$ and $j$)}}$$ 
Then $A_{n}[i,j]=1$ if $p_{ij}\!>\!threshold$ for a chosen 
threshold level. 
\end{enumerate}

Since N-gram and LDA  do not directly allow us to estimate $P(H_1)$, we 
use a Kolmogorov-Smirnov test on the words distribution to perform an 
overall hypothesis test between populations for the N-gram model, and a 
chi-square test for topic assignments across populations for LDA.

\subsection{Hyperparameters tunning}

 DD and our method require setting the number of clusters $H$, the 
 dimensionality of the low-rank factorization $R$, the Dirichlet 
 concentration parameters $\alpha$ for $\bm{\beta}$, and the prior 
 probability of $P(H_1)$. For our experiments, we fixed $H=15$, $R=10$, 
 $\alpha = 1/H$ and $P(H_1) = 0.5$.  
We found that $H=15$ was more than enough clusters 
for all instances, larger numbers resulting in empty clusters.  
 Most of our experiments focus on 
 settings with count-valued weights, and in the case of word 
 co-occurrences, the weight between words $i$ and $j$ is bounded by the 
 smaller of the number of occurrences of the two words $i$ and $j$. For 
 this setting, we therefore used the logit link and the binomial likelihood. 
On the other hand,  the Brain dataset has count-valued weights with 
unbounded support, and we used Exponential link and the Poisson likelihood.
Our results were fairly robust to hyperparameters settings. 
For our MCMC algorithm, we observed good mixing properties, and used 
$1300$ Gibbs samples with an extra $200$ burn-in samples. 

For LDA and N-Gram, we used settings following implementations from 
\cite{rehurek_lrec} and  \cite{ngrampkg}, respectively.   

\subsection{Results}

\vspace{1mm}
\noindent \textbf{Type-I error and statistical power:}
Type-I errors or false positives arise when a model incorrectly marks 
two populations as different when actually the null hypothesis is true, 
i.e., $P(H_1| H_1 \text{is false})$.  Ideally, type-I error rates 
{ should be } $0.05$ or less. 
Statistical power shows if the models can correctly determine when the 
populations are different, and $P(H_1 | H_1 \text{ is  true})$ should be 
close to $1$. Measuring these quantities requires access to ground 
truth. For the Brazil dataset, the disparity of political tendencies 
between opposition and proposition is clear enough that we treat it as ground truth (For the other real datasets, we do not have ground truth
available).

We consider four sample sizes for the synthetic data: $50$, $100$, $200$ 
and $300$. For the twitter data, we consider three sample sizes: $105 \:
(\sim\!\!5\%)$, $420 \: (\sim\!\!20\%)$ and $1049\: (\sim\!\!50\%)$. For DD which 
requires thresholding, we used four different threshold-levels, $0$, 
$25\%$, $50\%$ and $75\%$. For all methods, we compute $P(H_1| -)$ under 
different settings. In order to estimate variance $P(H_1| -)$, we generated 
20 datasets for each sample size.

Figure \ref{fig:syn_power} presents the Type-I error (i.e., $P(H_1| H_1 \text{is false})$) and power curves (i.e., $P(H_1| H_1 \text{is true})$) for 
increasing sample sizes for each method for synthetic data (left) and real world data (right). Each data point shown is the mean of the 20 trials for the respective sample size, we also present the range of $5\%$ to $95\%$ percentiles. We see that NC-F has the best overall performance both when $H_1$ is true and for $H_1$ false in the homogeneous scenario. N-gram has the worst performance overall so we do not consider this baseline in the real data. 
LDA have a good overall power, however it performs poorly as far as 
Type I error goes, with the largest variance. Further, LDA is not able to 
capture probabilistic structure underlying the data (we discuss this 
later). DD's performance is not so good for Type I error in the synthetic data, but it is good overall for power. This is not 
the case for the real data though, where power is poor and varies 
significantly with threshold. This sensitivity to threshold-level  
confirms the original motivation for this work.  NC-M outperformed all 
other methods for the heterogeneous data, and NC-F was the second best 
even though it does not account for heterogeneity. This is due to the 
fact that our method accounts for the weight distributions, and thus can 
handle overdispersed counts relatively well.

\begin{figure}[htbp]
\hspace{-8mm}
  \includegraphics[width=0.53\textwidth]{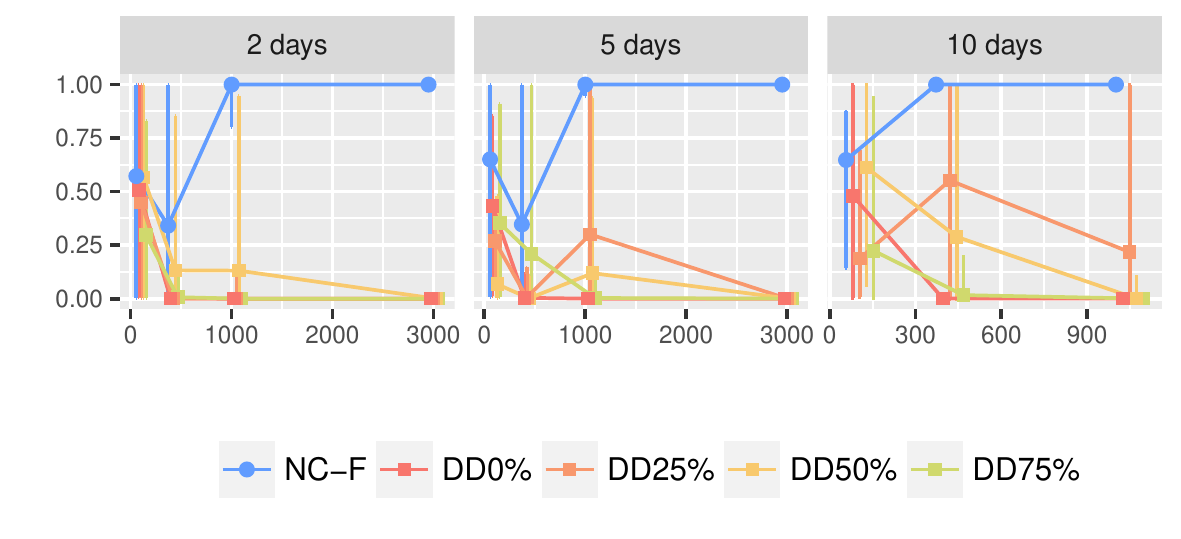}
  
  \caption{Power curves of NC-F and DD (multiple threshold levels) for increasing proportion of active users}
  \label{fig:power_active_sample}
\vspace{2mm}
\end{figure}

We also investigate how the number of active users in the dataset affects the overall power performance. First, we define ``activeness'' as a function of number of days a user tweeted. We created datasets restricting to users with at least $2$, $5$ and $10$ days of tweets. Thus, we assume that in the dataset with users having at least $10$ days of tweets, there are a significantly larger number of active users than in dataset with a minimum of $2$ days of tweets. Note that we are assessing only statistical power here, therefore we are only looking to the case that $H_1$ is true, i.e., ``PropositionVsOpposition''. Figure \ref{fig:power_active_sample} shows the power curve for each these datasets. As expected, our method needs fewer observations to find statistical difference between populations with more active users in the dataset, in other words NC-F has the ability to differentiate between populations easier as the proportion of active users increases. DD does not improve with larger sample size which suggests DD loses the ability to determine whether populations differ in a higher heterogeneous setting. 

\vspace{1mm}
\noindent \textbf{Population-level hypothesis test:}
In the previous results, we had a glimpse of the sensibility of threshold choice in terms of decision making on the hypothesis testing procedure. Here, we aim to analyze this further. We estimate the posterior probability of $H_1$ for \textit{all observations} of all the datasets.  We compare our results with DD for 10 different threshold levels $(0,10\%,20\%,...,90\%)$. Note that NC-F and NC-M do not vary with threshold level. Here, in addition to the Twitter data, we include results testing the fMRI dataset---comparing populations based on the creative index (CCI). \cite{durante2016} found a significant difference in Brain connectivity between non creative individuals ($CCI\le90$) vs creative subjects ($CCI\ge111$). In their tests, the graphs were thresholded at $0$.

\begin{figure}[h]
\centering
  \includegraphics[width=0.45\textwidth]{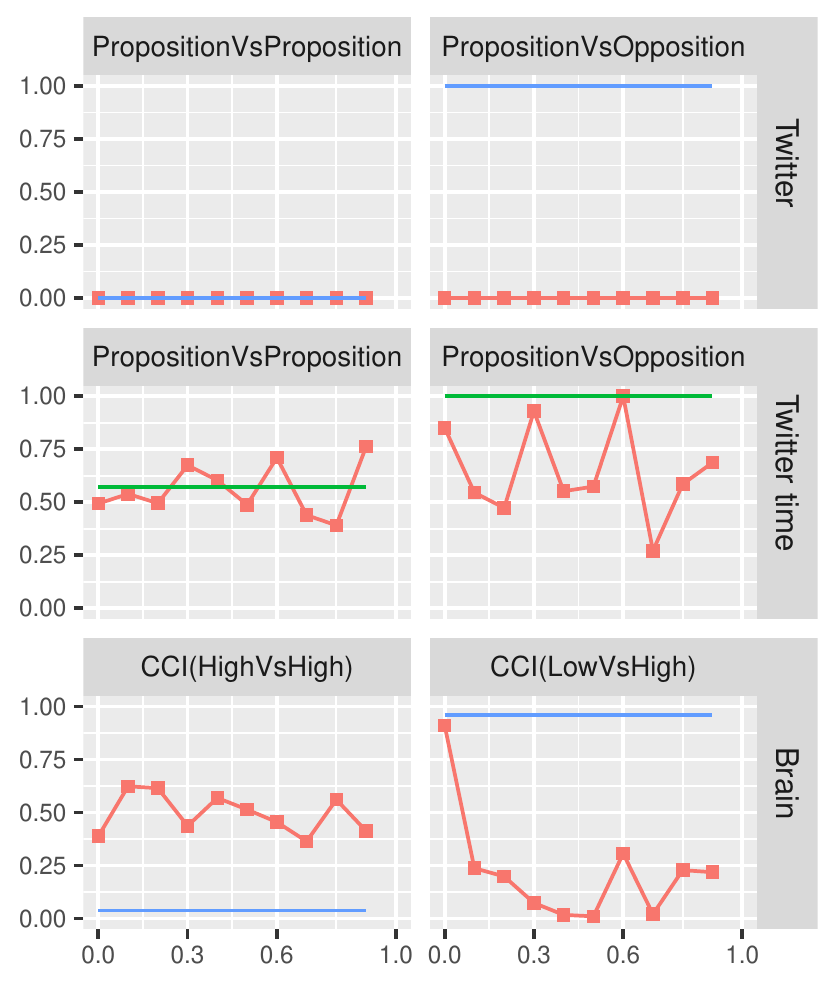}
  \caption{$P(H_1 | - )$ across threshold levels for three datasets for when $H_1$ is false (left column) and $H_1$ is true (right column). NC-F and NC-M are represented as blue and green lines, respectively.}
  \label{fig:ph1_all}
\vspace{2mm}
\end{figure}

Figure \ref{fig:ph1_all} shows DD represented as red solid squares, NC-F and NC-M as blue and green lines, respectively.  Again, DD exhibits sensitivity to the threshold choice making inferences unreliable. For instance, if we consider testing whether populations Proposition and Opposition are significantly different and use a 60\% threshold for Twitter time, we would reject the null, since the posterior of $P(H_1)\approx 1$. However, if we slightly change the threshold level to $50\%$, $P(H_1)\approx 0.5$  meaning that there is not enough evidence to support that they are statistically different and we accept the null. The same behavior can be seen on the fMRI dataset where the threshold at $0$ is statistical significant for \textit{LowVsHigh}, however it is not for any other threshold. Overall, we found that our methods NC-M and NC-F are more reliable, since they avoid the need for practitioners to make sensitive preprocessing choices.

\begin{figure}[h]
\hspace{-8mm}
\includegraphics[width=0.54\textwidth]{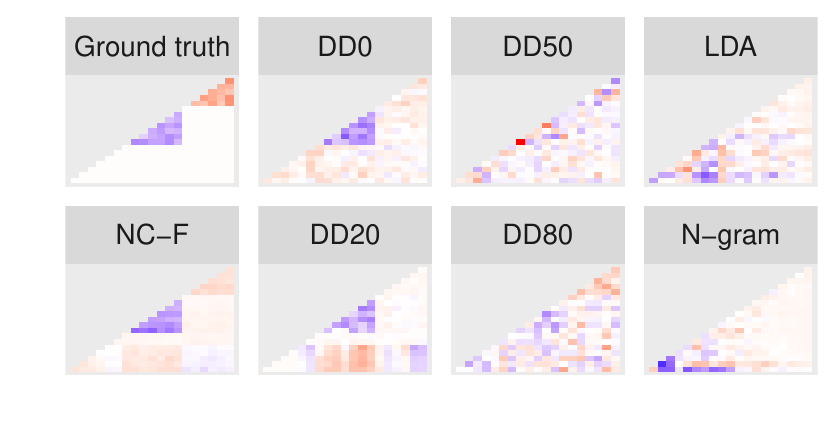}
 \vspace{-6mm}
  \caption{Edge-specific probabilities difference}
\label{edgeprob}
\vspace{-4mm}
\end{figure}

\vspace{1mm}
\noindent \textbf{Edge-specific level hypothesis test:}
\label{sec:edge_results}
Another important task is that of retrieving the structure of the co-occurrences probabilities. For better visualization, we generated a version of the synthetic homogeneous with 20 nodes, and we look at differences between true and predicted edge probability matrices for both populations, i.e.\ we compute the estimated difference $\bar{\bm{\theta}}_1 - \bar{\bm{\theta}}_2$ for each model and compared with the ground truth. In  Figure~\ref{edgeprob}, we see that our proposed framework accurately recovers the structure of
the ground truth. The DD0 also retrieves the structure of population 1, 
however it performs poorly for population 2. This is related to the
sensitivity of results to the threshold-level, suggesting this needs to be chosen carefully across different scenarios. Our models NC-F and NC-M both do not require such hand-tuning, and further exploit values of the pre-thresholded counts for more accurate inference. Unsurprisingly, all the other models fail to correctly learn the structure used to simulate the data. 

\begin{figure}[h]
\centering
\begin{subfigure}{0.2\textwidth}
\hspace{-1.1cm}
\centering
\includegraphics[trim={4mm 0 0 0},clip]{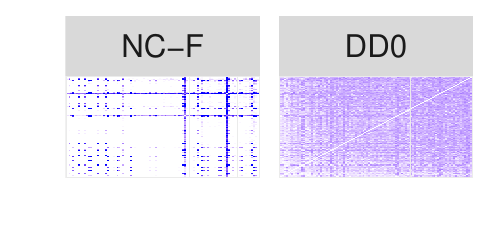} 
\end{subfigure}
\begin{subfigure}{0.2\textwidth}
\centering
\hspace{-1cm}
\includegraphics[trim={4mm 0 0 0},clip]{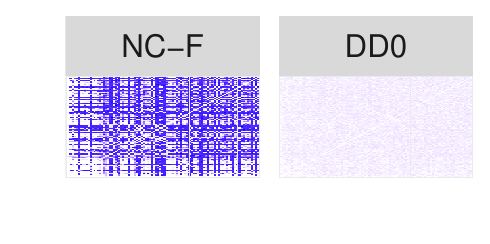} 
\end{subfigure}
 \vspace{-2mm}
\caption{Edge-specific tests for Twitter (left) and Instagram (right).} 
\label{lochypbra}
\vspace{2mm}
\end{figure}

For the Twitter and Instagram datasets, we looked at each edge, and 
identified those that are different using a $0.1$ significance level. 
Figure \ref{lochypbra} shows that the NC-F model was able to capture a clear pattern of significantly different use of words among populations for the Brazil dataset, as opposed to the other modeling schemes which look almost random. For the Instagram drug data, the edge-specific hypothesis testing matrix structure is much more significant compared with the Twitter case. One reason for this is that there is a group of Ritalin users that are German, and their words differ from others.

\begin{figure}[!h]
\vspace{3mm}
 \hspace{-2mm}
\centering
\includegraphics[width=.50\textwidth]{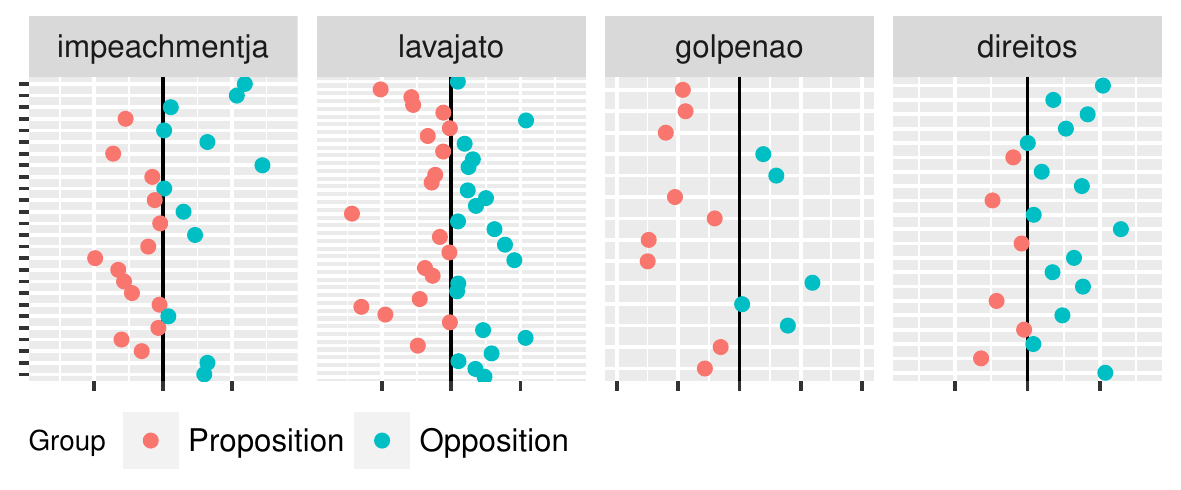} 
\caption{Swing words: differences in edge probabilities for example high frequency terms.}
\label{swingwords}
\vspace{3mm}
\end{figure}

\vspace{1mm}
\noindent \textbf{Exploratory analysis:}
Here, we show some additional insights that our methods are capable of capturing. One interesting fact of the Brazil political scenario is that many high frequency words were extensively used across both populations, examples being ``impeachmentja'' (impeachment now), ``lavajato'' (carwash), ``golpenao'' (no coup), ``direitos'' (rights). 
However, using the  probability structure $\bar{\bm{\theta}}$ estimated from our 
framework, we can make some interesting insights about how the two sides frame the issues differently. Figure \ref{swingwords} plots the difference of the link probability for each high frequency word used in conjunction (co-occurrence) with all other words, across the two sides---if the value is larger than zero then it is a `proposition expression', otherwise it is an  `opposition expression'. For instance, ``lavajato'' is the name of the investigation and if it is used with ``motivo'' (motive) is a proposition statement where if it is used with ``luta'' (fight) then it is a clear opposition one. From Figure \ref{swingwords} it is clear that the two sides use sets of words (e.g., phrases) quite differently. 

\begin{figure}[h]
\centering
  \includegraphics[width=.45\textwidth]{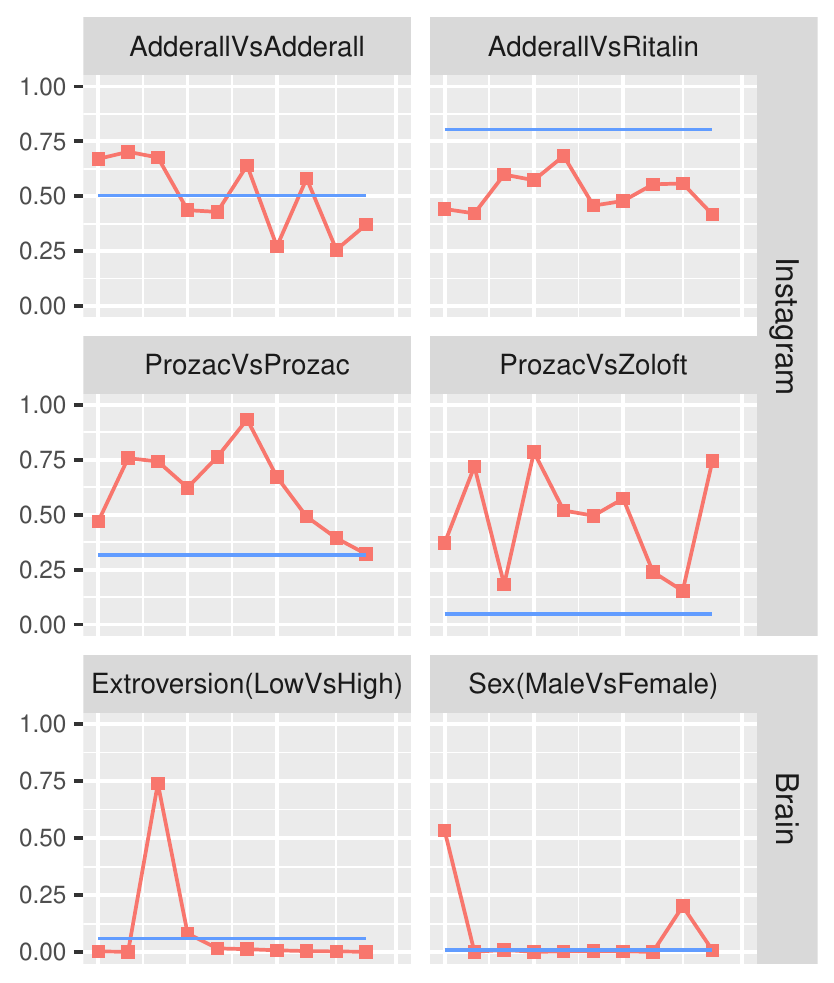}
  \caption{$P(H_1 | - )$ across threshold levels for Instagram and fMRI datasets.}
  \label{fig:ph1_all_explo}
\vspace{-4mm}
\end{figure}

Figure \ref{fig:ph1_all_explo} presents additional results for the Instagram and fMRI datasets. In this case, we look again to the behavior of $P(H_1|-)$ across threshold levels. It is important to highlight that we do not have a ground truth information to compare our findings with, however it is an additional set of results to  explore the assessment of significant difference between two populations, and to note the lack of robustness of DD wrt threshold choice. 

\section{Conclusion}
\label{sec:con}

This paper presents the first steps towards routine and systematic 
hypothesis testing on populations of weighted networks. Our statistical 
framework applies both to settings where entities from each population 
have single graphs associated with them, as well as settings where 
each entity has associated a set of graphs (we call these without and with 
within-population heterogeneity). 
Through a flexible and general clustering mechanism for replicated 
weighted networks, our framework offers a powerful and accurate 
hypothesis testing at three levels: population-level, entity-specific and  
edge-specific. 
We applied our model to study communication behavior on real social 
media data (Instagram and Twitter), as well as for brain connectivity 
data.  We saw that by not relying on a a user-specified threshold, our 
proposed method offers robustness over the methodology 
of~\cite{durante2016}, besides outperforming other baselines like LDA, 
N-gram language models.
\bibliographystyle{IEEEtran}
\newpage
\bibliography{mybib}
\section{Appendix}

\subsection{Synthetic data details}
\label{sec:struc}

\begin{figure}[tb]
  \centering
  \includegraphics[width=0.45\textwidth]{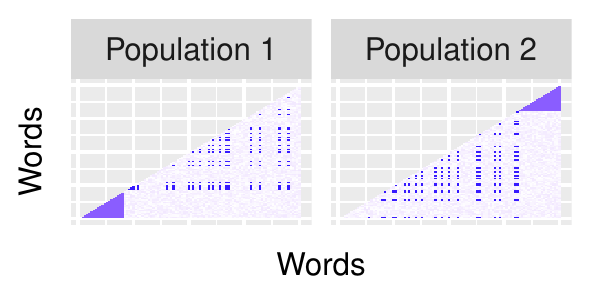}
  \caption{Structure used to simulate homogeneous data }
  \label{struc3}
\end{figure}

\begin{figure}
\centering
\includegraphics[width=0.45\textwidth]{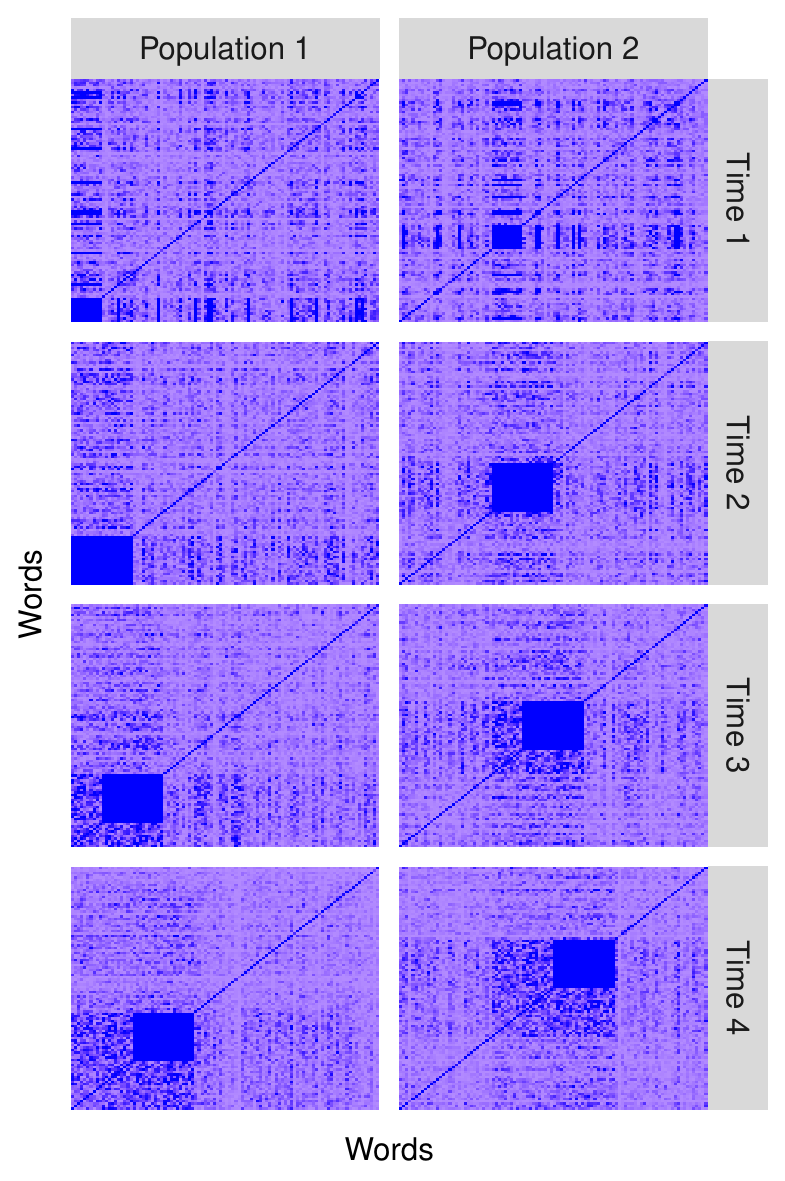} 
\caption{Structure used to simulate heterogeneous data }
\label{struc_time}
\end{figure}

The synthetic data was generated to have the same construction of the Twitter dataset set, i.e. a set word co-occurrences networks. In this sense, each node is a word and each edge is a co-occurrence of two words. Moreover, each pair of words can co-occur at most the minimum occurrence of each individual word. In other words, say words $i$ and $j$ occurred $x$ and $y$ times, respectively, then the co-occurrence of words $i$ and $j$ is at most $\min(x,y)$. Hence, in order to generate graphs in this setting, we need to have the individual occurrences of all the words and probability structure for the co-occurrences. Since  Zipf's law  \cite{zipf} , or discrete Pareto, is almost always used to describe words frequencies, we used a multivariate Zipf generating process \cite{paretomulti2002} to generate the individual counts. Equation \ref{jeh} shows how to generate multivariate Zipf's values, $\bm{X}_n \in \mathbb{R}^{V}$ is the vector with all the individual counts of $V$ words for entity $n$. In our case, we considered the standard Zipf where $\lambda = 1$.
\begin{equation}
\begin{aligned}[c]
p &\sim Beta(1,\lambda)\\
\bm{X}_n|p &\stackrel{ind}{\sim} Geo(p)
\end{aligned}
\Rightarrow
\begin{aligned}[c]
\bm{X}_n &\sim M^{(m)}Zipf(IV)(0,\lambda,1,1)
\end{aligned}
\label{jeh}
\end{equation}
The structures were arbitrarily chosen to have a clear difference across populations. Figure \ref{struc3} shows the structure used to simulate the homogeneous data used on the experiments section and Figure \ref{struc_time} shows the structure used to simulate for the heterogeneous data. Given the individual counts and the probability of each edge, we simulate edge count using Binomial distribution. Formally, $A_{nij} \sim Bin\left(\min\left({X_{ni},X_{nj}}\right),\bm{\theta}_{ij}\right)$, where $A_{nij}$ is the co-occurrence of words $i$ and $j$, $X_{ni}$ and $X_{nj}$ are their individual counts and $\bm{\theta}_{ij}$ is the probability of $i$ and $j$ co-occur once. 

\end{document}